\documentclass[10pt,twocolumn,letterpaper]{article}

\usepackage{cvpr}
\usepackage{times}
\usepackage{epsfig}
\usepackage{graphicx}
\usepackage{amsmath}
\usepackage{amssymb}

\usepackage{booktabs} %
\usepackage{xparse}   %
\NewDocumentCommand{\rot}{O{45} O{1em} m}{\makebox[#2][l]{\rotatebox{#1}{#3}}}%

\usepackage[pagebackref=true,breaklinks=true,letterpaper=true,colorlinks,bookmarks=false]{hyperref}

\cvprfinalcopy %

\def\cvprPaperID{7022} %

\ifcvprfinal\fi
\begin{document}

\title{FastDraw: Addressing the Long Tail of Lane Detection\\ by Adapting a Sequential Prediction Network}

\author{Jonah Philion \\
\href{http://isee.ai}{ISEE.AI}\\
{\tt\small jonahphilion@isee.ai}
}

\maketitle
\thispagestyle{empty}

\begin{abstract}
   The search for predictive models that generalize to the long tail of sensor inputs is the central difficulty when developing data-driven models for autonomous vehicles. In this paper, we use lane detection to study modeling and training techniques that yield better performance on real world test drives. On the modeling side, we introduce a novel fully convolutional model of lane detection that learns to decode lane structures instead of delegating structure inference to post-processing. In contrast to previous works, our convolutional decoder is able to represent an arbitrary number of lanes per image, preserves the polyline representation of lanes without reducing lanes to polynomials, and draws lanes iteratively without requiring the computational and temporal complexity of recurrent neural networks. Because our model includes an estimate of the joint distribution of neighboring pixels belonging to the same lane, our formulation includes a natural and computationally cheap definition of uncertainty. On the training side, we demonstrate a simple yet effective approach to adapt the model to new environments using unsupervised style transfer. By training FastDraw to make predictions of lane structure that are invariant to low-level stylistic differences between images, we achieve strong performance at test time in weather and lighting conditions that deviate substantially from those of the annotated datasets that are publicly available. We quantitatively evaluate our approach on the CVPR 2017 Tusimple lane marking challenge, difficult CULane datasets \cite{culane}, and a small labeled dataset of our own and achieve competitive accuracy while running at 90 FPS.
\end{abstract}

\section{Introduction}

\begin{figure}[t]
\begin{center}
   \includegraphics[width=0.95\linewidth]{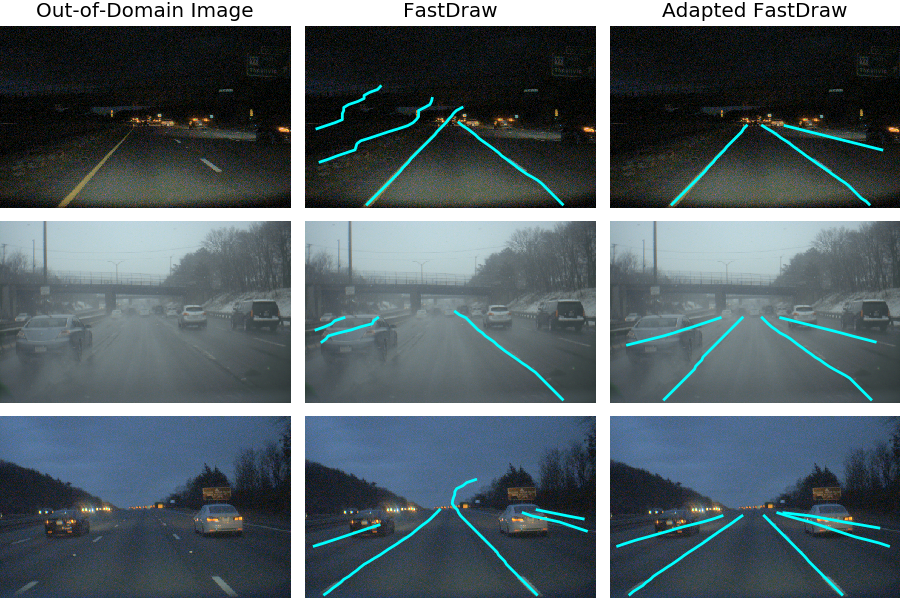}
\end{center}
   \caption{Best viewed in color. We train a novel convolutional lane detection network on a public dataset of labeled sunny California highways. Deploying the model in conditions far from the training set distribution (left) leads to poor performance (middle). Leveraging unsupervised style transfer to train FastDraw to be invariant to low-level texture differences leads to robust lane detection (right).}
\label{fig:brief}
\end{figure}

\begin{figure*}
\begin{center}
\includegraphics[width=0.9\linewidth]{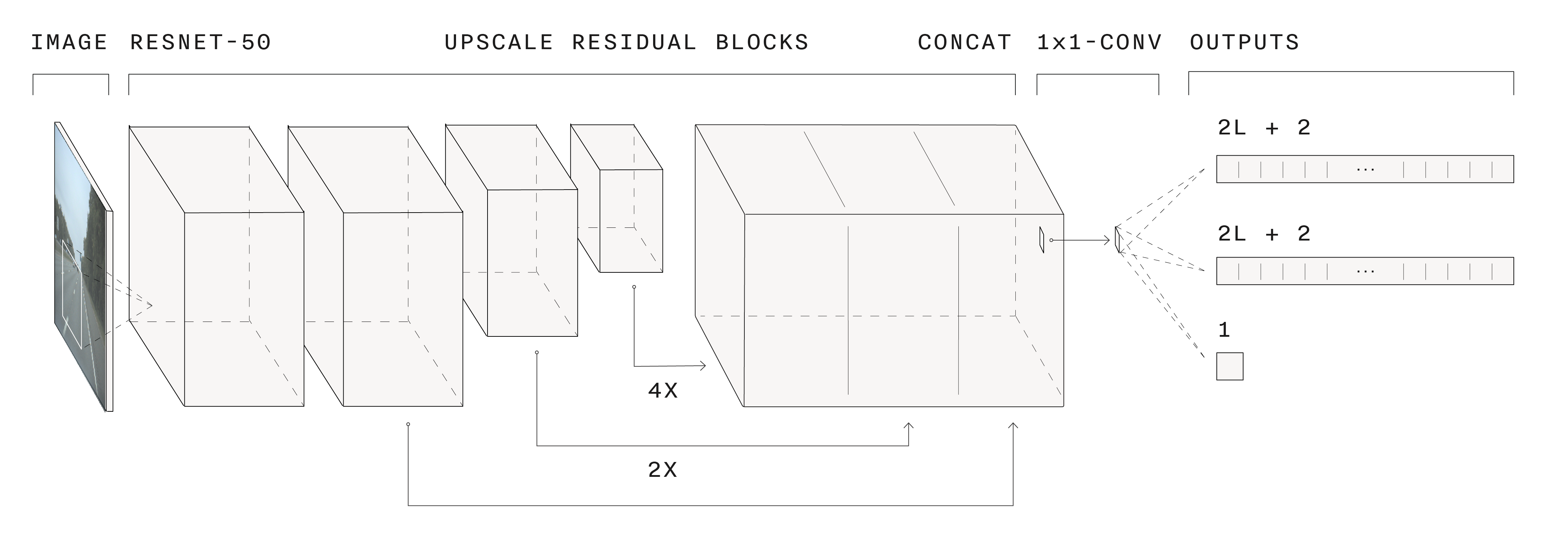}
\end{center}
   \caption{We use a CNN to extract a semantic representation of the input image. This representation is decoded by three separate shallow convolutional heads: a binary segmentation of pixels that belong to a lane ($p_{h,w,0}$), and a categorical distribution over the pixels within $L$ pixels of the current pixel in the rows above and below ($p_{h,w,1}$ and $p_{h,w,-1}$ respectively). Because we include an \texttt{end} token in the categorical distribution to train to the network to predict endpoints, the categorical distributions are $2L + 1 + 1=2L+2$ dimensional.}
\label{fig:model}
\end{figure*}

Previous models of lane detection generally follow the following three-step template. First, the likelihood that each pixel is part of a lane is estimated. Second, pixels that clear a certain threshold probability $p_{min}$ of being part of a lane are collected. Lastly, these pixels are clustered, for instance with RANSAC, into individual lanes.\\
\indent Because the second and third steps in which road structure is inferred from a point cloud of candidate pixels are in general not differentiable, the performance of models of lane detection that follow this template is limited by the performance of the initial segmentation. We propose a new approach to lane detection in which the network performs the bulk of the decoding, thereby eliminating the need for hyper-parameters in post-processing. Our model ``draws" lanes in the sense that the network is trained to predict the local lane shape at each pixel. At test time, we decode the global lane by following the local contours as predicted by the CNN.\\
\indent A variety of applications benefit from robust lane detection algorithms that can perform in the wild. If the detector is iterative, the detector can be used as an interactive annotation tool which can be used to decrease the cost of building high definition maps \cite{uberlane,polyrnnpp}. For level 5 systems that depend on high definition maps, online lane detection is a useful localization signal. Level 2 systems that are not equipped to handle the computational load required of high definition maps depend on models of lane detection equipped with principled methods of determining when to notify the driver that the lane detection is uncertain. In pursuit of solutions for these applications, we identify three characteristics that a lane detection module should possess.\\
\indent First, the lane detection algorithm must be able to represent any number of lanes of any length. Whereas variability in the number of instances of an object in an image is an aspect of any kind detection problem, variability in the dimensionality of a single instance is a more unique to the lane detection problem; unlike bounding boxes which have a precise encoding of fixed dimensionality, lane segments can be arbitrary length. Solutions that reduce lanes to a constant dimensionality - such as by fitting them with polynomials - lose accuracy on tight curves where accurate lane detection or localization is important for safe driving.\\
\indent Second, the detection algorithm must run in real-time. Therefore, although there is variability in the number and size of lanes in an image, whatever recursion used to identify and draw these lanes must be fast. Solutions to the variable dimensionality problem that involve recurrent cells \cite{lstm} or attention \cite{attn} are therefore a last resort.\\
\indent Finally, the detection algorithm must be able to adapt quickly to new scenes. Sensors such as cameras and lidar that are used in self-driving carry with them a long tail in the distribution of their outputs. A lane detection algorithm should be able to adapt to new domains in a scalable way.\\
\indent We present an approach which addresses these problems and is competitive with other contemporary lane detection algorithms. Our contributions are
\vspace{-5pt}
\begin{itemize}
    \item A lane detection model that integrates the decoding step directly into the network.  Our network is autoregressive and therefore comes equipped with a natural definition of uncertainty. Because decoding is largely carried out by the convolutional backbone, we are able to optimize the network to run at 90 frames per second on a GTX 1080. The convolutional nature of FastDraw makes it ideal for multi-task learning \cite{multi-task} or as an auxiliary loss \cite{waymo}. 
    \item A simple but effective approach to adapt our model to handle images that are far from the distribution of images for which we have public annotations. Qualititative results are shown in Figure \ref{fig:brief} and Figure \ref{fig:showoff}. While style transfer has been used extensively to adapt the output distribution of simulators to better match reality \cite{munitimage}, we use style transfer to adapt the distribution of images from publicly available annotated datasets to better match corner case weather and environmental conditions.
\end{itemize}

\section{Related Work}
\textbf{Lane Detection} Models of lane detection generally involve extracting lane marking features from input images followed by clustering for post-processing. On well-maintained roads, algorithms using hand-crafted features work well \cite{DBLP:journals/corr/Aly14,7217838}. Recent approaches such as those that achieve top scores on the 2017 Tusimple Lane Detection Challenge seek to learn these hand-crafted features in a more end-to-end manner using convolutional neural networks. To avoid clustering, treating left-left lane, left lane, right lane, and right-right lane as channels of the segmentation has been explored \cite{culane}. Projecting pixels onto a ground plane via a learned homography is a strong approach for regularizing the curve fitting of individual lanes \cite{end-to-end}. Research into improving the initial segmentation has been successful although results are sensitive to the heuristics used during post-processing \cite{elgan}. Recent work has improved segmentation by incorporating a history of images instead of exclusively conditioning on the current frame \cite{history}.\\
\indent Lane detection is not isolated to dashcam imagery. Models that detect lanes in dashcams can in general be adapted to detect lanes in lidar point-clouds, open street maps, and satellite imagery \cite{uberlane, maplanes, satellite-lane}. The success of semantic segmentation based approaches to lane detection has benefited tremendously from rapid growth in architectures that empirically perform well on dense segmentation tasks \cite{segnet, retinanet, feature-pyramid}. Implicitly, some end-to-end driving systems have been shown to develop a representation of lanes without the need for annotated data \cite{driveinaday, DBLP:journals/corr/BojarskiTDFFGJM16}.\\
\indent \textbf{Style Transfer} Adversarial loss has enabled rapid improvements in a wide range of supervised and unsupervised tasks. Pix2Pix \cite{pix2pix} was the first to demonstrate success in style translation tasks on images. Unsupervised style transfer for images \cite{unit,cyclegan} and unsupervised machine translation \cite{lampleunmt,artetxeunmt} use back-translation as a proxy for supervision. While models of image-to-image translation have largely been deterministic \cite{unit, cyclegan}, MUNIT \cite{munit} extends these models to generate a distribution of possible image translations. In this work, we incorporate images generated by MUNIT translated from a public dataset to match the style of our own cameras. We choose MUNIT for this purpose because it is unsupervised and generative. Previous work has used data from GTA-V to train object detectors that operate on lidar point clouds \cite{wu2018squeezesegv2}. Parallel work has shown that synthetic images stylized by MUNIT can improve object detection and semantic segmentation \cite{munitimage}. We seek to leverage human annotated datasets instead of simulators as seeds for generating pseudo training examples of difficult environmental conditions.\\
\indent \textbf{Drawing} We take inspiration from work in other domains where targets are less structured than bounding boxes such as human pose estimation \cite{Cao_2017} and automated object annotation \cite{polyrnnpp}. In human pose estimation, the problem of clustering joints into human poses has been solved by inferring slope fields between body parts that belong to the same human \cite{Cao_2017}. Similarly, we construct a decoder which predicts which pixels are part of the same lane in addition to a segmentation of lanes. In Polygon-RNN \cite{polyrnn} and Sketch-RNN \cite{sketchrnn}, outlines of objects are inferred by iteratively drawing bounding polygons. We follow a similar model of learned decoding while simplifying the recurrence due to the relative simplicity of the lane detection task and need for realtime performance.

\section{Model}

Our model maximizes the likelihood of polylines instead of purely predicting per-pixel likelihoods. In doing so, we avoid the need for heuristic-based clustering post-processing steps \cite{DBLP:journals/corr/Aly14}. In this section, we describe how we derive our loss, how we decode lanes from the model, and how we train the network to be robust to its own errors when conditioning on its own predictions at test time.
\subsection{Lane Representation}
\indent In the most general case, lane annotations are curves $\gamma: [0,1] \rightarrow \mathbb{R}^2$. In order to control the orientation of the lanes, we assume that lane annotations can be written as a function of the vertical axis of the image. A lane annotation $\textbf{y}$ therefore is represented by a sequence of $\{height,width\}$ pixel coordinates $\textbf{y} = \{y_1,...,y_n \} = \{\{h_1,w_1\},...,\{h_n,w_n\}\}$ where $h_{i+1} - h_{i} = 1$.

Given an image $\textbf{x} \in \mathbb{R}^{3 \times H  \times W}$, the joint probability $p(\textbf{y} | \textbf{x})$ can be factored
\begin{equation}
    p(\textbf{y} | \textbf{x}) = p(y_1 | \textbf{x})\prod_{i=1}^{n-1} p(y_{i+1} | y_1,...,y_i, \textbf{x}).
\end{equation}
One choice to predict $ p(y_{i+1} | y_1,...,y_{i}, \textbf{x})$ would be to use a recurrent neural network \cite{uberlane, rnnlane}. Do decode quickly, we assume most of the dependency can be captured by conditioning only on the previous decoded coordinate
\begin{equation}
    p(\textbf{y} | \textbf{x}) \approx p(y_1 | \textbf{x})\prod_{i=1}^{n-1} p(y_{i+1} | y_i, \textbf{x}).
    \label{eqn:prob}
\end{equation}
Because we assume $h_{i+1} - h_i=1$, we can simplify
\begin{align}
    p(y_{i+1} | y_{i}, \textbf{x}) &= p(\Delta w_i | y_{i}, \textbf{x})\\
    \Delta w_i &= w_{i+1} - w_i .
\end{align}
Lane detection is then reduced to predicting a distribution over $dw/dh$ at every pixel in addition to the standard per-pixel likelihood. Decoding proceeds by choosing an initial pixel coordinate and integrating.\\
\indent To represent the distribution $p(\Delta w_i | y_i, \textbf{x})$, we could use a normal distribution and perform regression. However, in cases where the true distribution is multi-modal such as when lanes split, a regression output would cause the network to take the mean of the two paths instead of capturing the multimodality. Inspired by WaveNet \cite{wavenet}, we choose to make no assumptions about the shape of $p(\Delta w_i | y_i, \mathbf{x})$ and represent the pairwise distributions using categorical distributions with support $\Delta w \in \{ i \in \mathbb{Z} | -L \leq i \leq L \} \cup \{ \texttt{end} \}$ where $L$ is chosen large enough to be able to cover nearly-horizontal lanes and \texttt{end} is a stop token signaling the end of the lane. At each pixel $\{h,w\}$, our network predicts
\begin{itemize}
    \item $p_{h,w,0} := p(y_{h,w} = 1 | \textbf{x})$ - the probability that pixel $\{h, w\}$ is part of a lane.
    \item $p_{h,w,1} := p(\Delta w^{+1} | y_{h,w}=1, \textbf{x})$ - the categorical distribution over  pixels in the row \textbf{above} pixel $\{h,w\}$ within a distance $L$ that pixel $\{h+1,w+\Delta w^{+1}\}$ is part of the same lane as pixel $\{h,w\}$ or that pixel $\{h,w\}$ is the \textbf{top} pixel in the lane it is a part of.
    \item $p_{h,w,-1} := p(\Delta w^{-1} | y_{h,w}=1, \textbf{x})$ - the categorical distribution over pixels in the row \textbf{below} pixel $\{h,w\}$ within a distance $L$ that pixel $\{h-1,w+\Delta w^{-1}\}$ is part of the same lane as pixel $\{h,w\}$ or that pixel $\{h,w\}$ is the \textbf{bottom} pixel in the lane it is a part of.
\end{itemize}
Given these distributions, we can quickly decode a full lane segment given any initial point on the lane. Given some initial position $h_0,w_0$ on lane $\textbf{y}$, we follow the greedy recursion
\begin{align}
    \textbf{y}(h_0) &= w_0\\
    \textbf{y}(h+d) &= \textbf{y}(h) + \Delta w^{d}\\
    \Delta w^{d} &= - L + \text{argmax} p_{h,\textbf{y}(h),d}
\end{align}
where $d \in \{-1,1\}$ during the downwards and upwards drawing stages. Note that we can choose any $y_i \in \textbf{y}$ as $h_0,w_0$ as long as we concatenate the results from the upwards and downwards trajectories. We stop decoding when argmax returns the \texttt{end} token.

\subsection{Architecture}
To extract a semantic representation of the input image, we repurpose Resnet50/18 \cite{DBLP:journals/corr/HeZRS15} for semantic segmentation. The architecture is shown in Figure \ref{fig:model}. We use two skip connections to upscale and concatenate features at a variety of scales. All three network heads are parameterized by two layer CNNs with kernel size 3. In all experiments, we initialize with Resnets pretrained on Imagenet \cite{pytorch}.

\subsection{Loss}

\begin{figure}[t]
\begin{center}
   \includegraphics[width=0.6\linewidth]{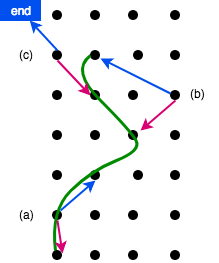}
\end{center}
   \caption{In addition to predicting per-pixel likelihoods, we train our network to output the pixels in the row above (blue) and below (purple) that are in the same lane as the current pixel. We also train pixels that are offset from annotated lanes to point back to the annotated lane (b). We include the \texttt{end} token in the categorical distribution to signal the termination of a lane (c). Given these predictions, we draw lanes by sampling an initial point then greedily following arrows up and down until we reach \texttt{end} in either direction and concatenating the two results. }
\label{fig:noisify}
\end{figure}

We minimize the negative log likelihood given by \eqref{eqn:prob}. Let $\theta$ represent the weights of our network, $\mathbf{x} \in \mathbb{R}^{3,H,W}$ an input image, $\mathbf{y} = \{\{h_1,w_1\},...,\{h_n,w_n\}\}$ a ground truth lane annotation such that $h_i - h_{i-1} = 1 $ and $\mathbf{y_m} \in \mathbb{R}^{1,H,W}$ a ground truth segmentation mask of the lane. The loss $L(\theta)$ is defined by
\begin{align}
    L_{mask}(\theta) &= -\log(p(\mathbf{y_m} | f_\theta(\mathbf{x}) ))\\
    \begin{split}
    L_{sequence}(\theta) &=\\
    -\sum_{d \in \{-1, 1\}}&\sum_{i=1}^{n} \log(p(\Delta w_i^d | f_\theta(\textbf{x})))
    \end{split}\\
    L(\theta) &= L_{mask}(\theta) + L_{sequence}(\theta)
\end{align}
Because the the task of binary segmentation and pairwise prediction have different uncertainties and scales, we dynamically weight these two objectives \cite{multi-task}. We incorporate a learned temperature $\sigma$ which is task specific to weigh our loss:
\begin{align}
\begin{split}
    L(\theta) =& \frac{1}{\sigma_{mask}^2} L_{mask}(\theta) + \frac{1}{\sigma_{sequence}^2} L_{sequence}(\theta) \\
    &+ \log \sigma_{mask}^2 \sigma_{sequence}^2.
\end{split}
\label{eqn:big_l}
\end{align}
During training, we substitute $W = \log \sigma^2$ into \eqref{eqn:big_l} for numerical stability. Experiments in which we fixed $W$ resulted in similar performance to allowing $W$ to be learnable. However, we maintain the dynamically weighed loss for all results reported below to avoid tuning hyperparameters.

\subsection{Exposure Bias}
Because our model is autoregressive, training $p_{h,w,\pm1}$ by conditioning exclusively on ground truth annotations leads to drifting at test time \cite{exposurebias}. One way to combat this issue is to allow the network to act on its own proposed decoding and use reinforcement learning to train. The authors of \cite{polyrnnpp} take this approach using self-critical sequence training \cite{selfcriticalrl} and achieve good results.\\
\indent Although we experiment with reinforcement learning, we find that training the network to denoise lane annotations --- also known as augmenting datasets with ``synthesized perturbations" \cite{waymo} --- is significantly more sample efficient. This technique is visualized in Figure \ref{fig:noisify}. To each ground truth annotation $\textbf{y}$ we add gaussian noise and train the network to predict the same target as the pixels in $\textbf{y}$. We therefore generate training examples
\begin{align}
    s &\sim \left \lfloor{\mathcal{N}(0.5, \sigma)}\right \rfloor&\\
    w_i ' &= w_i + s&\\
    \Delta w_i^d &= w_{i+s} - w_i '
\end{align}
where $d \in \{-1, 1\}$. We tune $\sigma$ as a hyperparameter which is dependent on dataset and image size. We clamp all $\Delta w$ between $-L$ and $L$ and clamp all $w'$ between 0 and the width of the image.

\begin{figure}[t]
\begin{center}
   \includegraphics[width=0.95\linewidth]{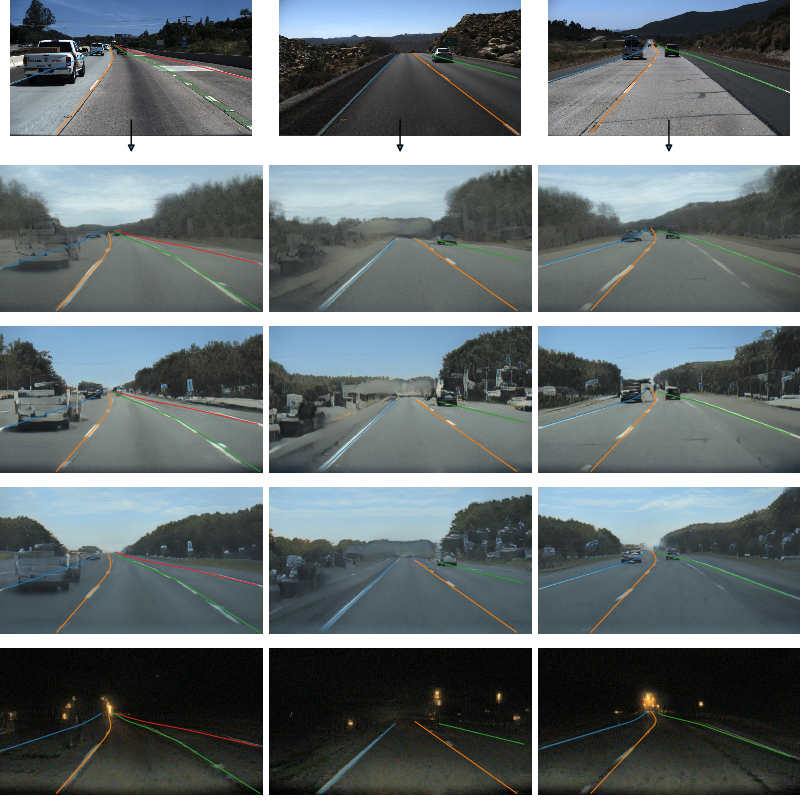}
\end{center}
   \caption{The top row shows three images $\textbf{x}_i$ from the Tusimple dataset and their annotations. The bottom four rows display samples from $G(\textbf{x}_i)$ with the adjusted Tusimple annotations overlaid. We use these additional training samples to bias the network towards shape instead of texture \cite{textureinvar}.}
\label{fig:augdata}
\end{figure}

\subsection{Adaptation}
The downside of data-driven approaches is that we have weak guarantees on performance once we evaluate the model on images far from the distribution of images that it trained on. For this purpose, we leverage the MUNIT framework \cite{munit} to translate images from public datasets with ground truth annotations into a distribution of images we acquire by driving through Massachusetts in a variety of weather and lighting conditions.\\
\indent To perform style transfer on the images in unordered datasets $D$ and $D'$, the CycleGAN framework \cite{cyclegan} trains an encoder-generator pair $E,G$ for each dataset $D$ and $D'$ such that $G(E(\mathbf{x})) \approx \mathbf{x}$ for $\mathbf{x} \sim D$ and difference between the distributions $y \sim G'(E(\mathbf{x}))$ and $y \sim D'$ is minimized, with analogous statements for $D'$. The MUNIT framework generalizes this model to include a style vector $s \sim \mathcal{N}(0, \mathbf{I})$ as input to the encoder $E$. Style translations are therefore distributions that can be sampled from instead of deterministic predictions.\\
\indent As shown in Figure \ref{fig:augdata}, we use MUNIT to augment our labeled training set with difficult training examples. Let $D = \{\mathbf{x_i}, \mathbf{y_i}\}$ be a dataset of images $\mathbf{x}_i$ and lane annotations $\mathbf{y}_i$ and $D' = \{\mathbf{x}_i\}$ a corpus of images without labels. Empirically, we find that style transfer preserves the geometric content of input images. We can therefore generate new training examples $\{\mathbf{x}', \mathbf{y}'\}$ by sampling from the distribution $D' \sim \{\mathbf{x}', \mathbf{y}'\}$ defined by
\begin{align}
    \mathbf{x}, \mathbf{y} &\sim D\\
    \textbf{x}' &\sim G'(E(\mathbf{x}, s))_{s \sim \mathcal{N}(0, \mathbf{I})}\\
    \textbf{y}' &= \mathbf{y}
\end{align}
Although representation of lanes around the world are location dependent, we theorize that the distribution of lane geometries is constant. Unsupervised style transfer allows us to adjust to different styles and weather conditions without the need for additional human annotation.

\section{Experiments}
We evaluate our lane detection model on the Tusimple Lane Marking Challenge and the CULane datasets \cite{culane}. The Tusimple dataset consists of 3626 annotated 1280x720 images taken from a dash camera as a car drives on California highways. The weather is exclusively overcast or sunny. We use the same training and validation split as EL-GAN \cite{elgan}. In the absence of a working public leaderboard, we report results exclusively on the validation set. We use the publicly available evaluation script to compute accuracy, false positive rate, and true positive rate.\\
\indent Second, we adopt the same hyperparameters determined while training on Tusimple and train our network on the challenging CULane dataset. CULane consists of 88880 training images, 9675 validation images, and 34680 test images in a diversity of weather conditions scenes. The test set includes environmental metadata about images such as if the image is crowded, does not have lane lines, or is tightly curved. We report evaluation metrics on each of these cases as is done by CULane \cite{culane}.\\
\indent Finally, to evaluate the effectiveness of our model to adapt to new scenes, we drive in Massachusetts in a diversity of weather conditions and record 10000 images of dash cam data. Using publicly available source code, we train MUNIT to translate between footage from the Tusimple training set and our own imagery, then sample 10000 images from the generator. We note that upon evaluating qualitatively the extent to which the original annotations match the generated images, the frame of the camera is transformed. We therefore scale and bias the height coordinates of the original annotations with a single scale and bias across all images to develop $D'$. 
\begin{equation}
    w_i^* = m w_i + b
\end{equation}
Samples from the pseudo training example generator are shown in Figure \ref{fig:augdata}. A speed comparison of FastDraw to other published models is shown in Table \ref{tab:performance_table}. For models trained on images of size $128 \times 256$ we use $L=6$ pixels and $\sigma=2$ pixels. For models trained on images of size $352 \times 640$ we use $L=16$ pixels and $\sigma=5$ pixels.

\section{Results}
\subsection{Tusimple}
We train FastDraw on the Tusimple dataset. We train for $7$ epochs in all experiments with batch size 4 using the Adam optimizer \cite{adam}. We initialize learning rate to 1.0e-4 and halve the learning rate every other epoch. To generate the candidates for initial points of lanes, we mask the center of the image and use DBSCAN from scikit-learn \cite{scikit} with $\epsilon=5$ pixels to cluster. Given these initial positions, we follow the predictions of FastDraw to draw the lane upwards and downwards.\\
\indent We compare our algorithm quantitatively against EL-GAN \cite{elgan}. EL-GAN improves lane detection over traditional binary segmentation by adding an additional adversarial loss to the output segmentation. The segmentations therefore approximate the distribution of possible labels. However, this approach still requires a heuristic decoder to convert the segmentation into structured lane objects. FastDraw is competitive with EL-GAN by all Tusimple metrics as shown in Table \ref{tab:tusimple}.\\

\begin{figure}[t]
\begin{center}
   \includegraphics[width=0.95\linewidth]{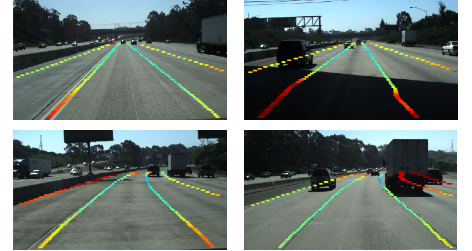}
\end{center}
   \caption{The standard deviation of the distribution predicted by FastDraw is plotted as error bars on a variety of images from Tusimple test set. Our color map is thresholded at a standard deviation of 0 and 9 pixels. We find that the network is accurately growing more uncertain in regions where the exact location of a lane is not well defined, for instance when the lane marking is wide, there are shadows, the lane is possibly a false positive, or the lane is occluded by other vehicles.}
\label{fig:uncertain}
\end{figure}

\begin{table}[t]
\begin{center}
\begin{tabular}{|l | c c c|}
\hline
Method & Acc (\%) & FP & FN \\
\hline\hline
EL-GAN (basic) & 93.3 & 0.061 & 0.104 \\
EL-GAN (basic++) & 94.9 & \textbf{0.059} & 0.067 \\
\hline
FastDraw Resnet18 & 94.9 & 0.061 & 0.047\\
FastDraw Resnet50 & 94.9 & \textbf{0.059} & 0.052\\
FastDraw Resnet50 (adapted) & \textbf{95.2} & 0.076 & \textbf{0.045}\\
\hline
\end{tabular}
\end{center}
\caption{We use Tusimple evaluation metrics to compare quantitatively with EL-GAN \cite{elgan}. FastDraw achieves comparable performance to EL-GAN with fewer layers. We note that while the accuracy of adapted FastDraw achieves high accuracy, it also has the highest false positive rate. We reason that the network learns a stronger prior over lane shape from $D'$, but the style segmentation does not always preserve the full number of lanes which results in the side of the road being falsely labeled a lane in the $D'$ dataset.}
\label{tab:tusimple}
\end{table}

\subsection{Uncertainty}
Because our network predicts a categorical distribution at each pixel with suppoert $\Delta w \in \{ i \in \mathbb{Z} | -L \leq i \leq L \} \cup \{ \texttt{end} \}$, we can quickly compute the standard deviation of this distribution conditioned on $\Delta w \neq \texttt{end}$ and interpret the result as errorbars in the width dimension. We find that the uncertainty in the network increases in occluded and shadowy conditions. Example images are shown in Figure \ref{fig:uncertain}. These estimates can be propagated through the self-driving stack to prevent reckless driving in high-uncertainty situations.
\subsection{Is the learned decoder different from a simple heuristic decoder?}
A simple way to decode binary lane segmentations is to start at an initial pixel, then choose the pixel in the row above the initial pixel with the highest probability of belonging to a lane. To show that our network is not following this decoding process, we calculate the frequency at which the network chooses pixels that agree with this simple heuristic based approach. The results are shown in Table \ref{tab:argmax}. We find that although the output of the two decoders are correlated, the learned decoder is in general distinct from the heuristic decoder.

\begin{table}[ht]
\begin{center}
\begin{tabular}{|l  c|}
\hline
Model & Frames-Per-Second \\
\hline\hline
Ours & \textbf{90.31} \\

H-Net [20] & 52.6\\

CULane [21] & 17.5\\

PolyLine-RNN [9] & 5.7\\

EL-GAN [7] & \textless 10\\
\hline
\end{tabular}
\end{center}
\caption{Because FastDraw requires little to no post-processing, the runtime is dominated by the forward pass of the CNN backbone which can be heavily optimized.}
\label{tab:performance_table}
\end{table}

\begin{figure*}
\begin{center}
\includegraphics[width=1.0\linewidth]{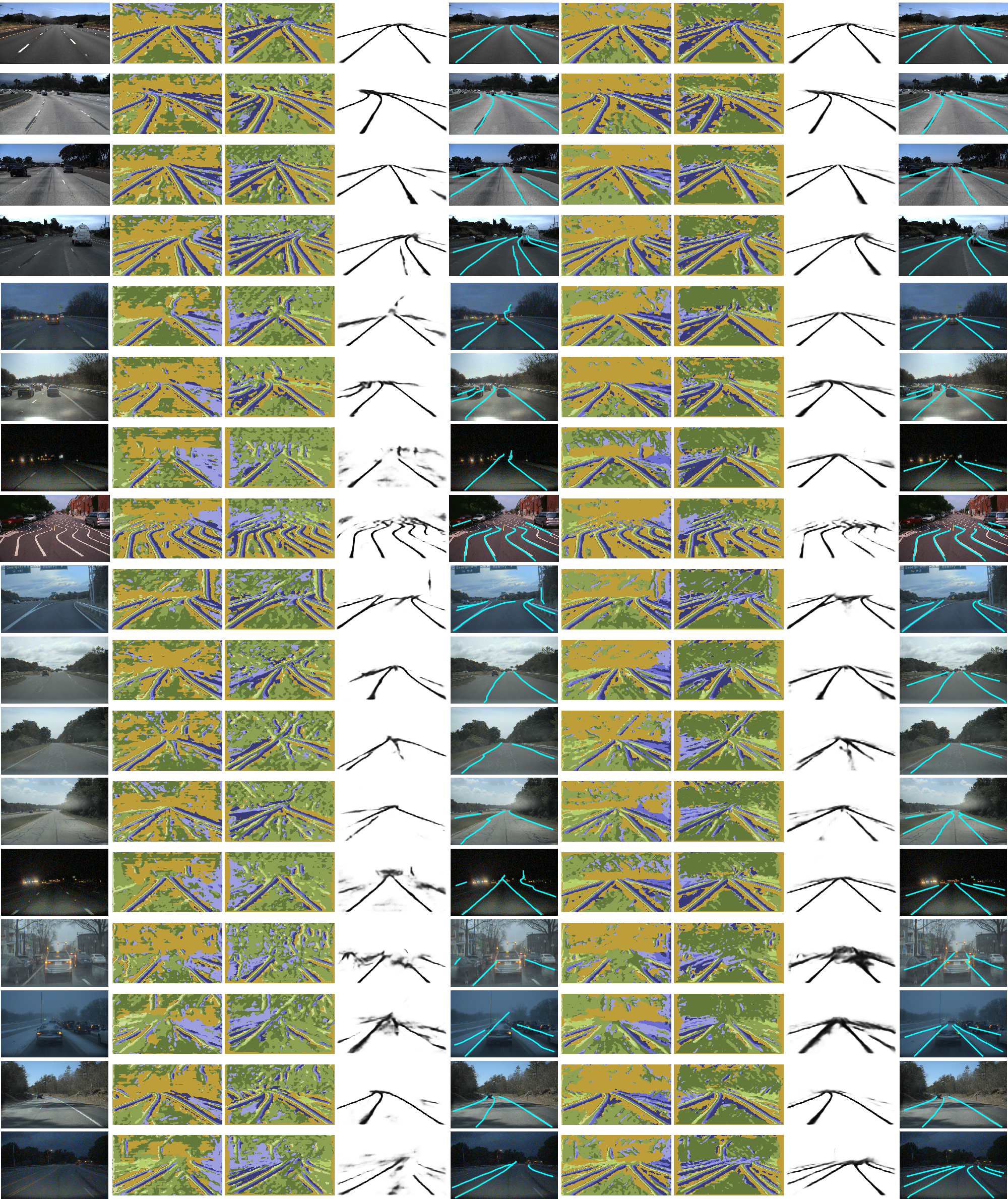}
\end{center}
   \caption{We demonstrate that our network can perform with high accuracy on the publicly annotated dataset we train on as well as images collected from a diversity of weather conditions and scenes in Massachusetts. Column 1 shows the ``lookup" head argmax$p_{h,w,1}$, column 2 shows the ``lookdown" head argmax$p_{h,w,-1}$, column 3 shows the per-pixel likelihood $p_{h,w,0}$, and column 4 shows the decoded lanes on top of the original image. Columns 4-8 show the analogous visual for adapted FastDraw.  Images in the top 4 rows come from the Tusimple test set. The rest come from our own driving data collected in the wild on Massachusetts highways. We find that the data augmentation trains the network to draw smoother curves, recognize the ends of lanes better, and find all lanes in the image. Even on Tusimple images, the annotation improves the per-pixel likelihood prediction. At splits, the network chooses chooses a mode of the distribution instead of taking the mean.}
\label{fig:showoff}
\end{figure*}

\begin{table}
\begin{center}
\begin{tabular}{|l|l|}
\hline
$| p_{h,w,1} - \text{argmax} p_{h+1 , w+\Delta w, 0} |$ & \% \\
\hline\hline
$<1$ & 12.6\\
$<3$ & 58.2\\
$<5$ & 87.1\\
\hline
\end{tabular}
\end{center}
\caption{For a series of decoded polylines, we calculate the distance between the pixel with maximum likelihood $p_{h\pm1,w,0}$ in the row above $\{h,w\}$ with the chosen $\text{argmax} p_{h,w,\pm1}$. We report the percent of the time that the distance is less than one, three, and five pixels. We find that the network is in general predicting values which agree with the $p_{h,w,0}$ predictions. Deviation from the naive decoder explains why the network still performs is able to perform well when the segmentation mask is noisy.}
\label{tab:argmax}
\end{table}

\subsection{CULane}
We train FastDraw on the full CULane training dataset \cite{culane}. We use identical hyperparameters to those determined on Tusimple with an exponential learning rate schedule with period $1000$ gradient steps and multiplier 0.95. FastDraw find the model achieves competitive performance with the CULane Resnet-50 baseline. IoU at an overlap of 0.5 as calculated by the open source evaluation provided by CULane is shown in Table \ref{tab:culane}. Of note is that FastDraw outperforms the baseline on curves by a wide margin as expected given that CULane assumes lanes will be well represented by cubic polynomials while FastDraw preserves the polyline representation of lanes.

\begin{table}
\begin{center}
\begin{tabular}{|l||cc|}
\hline
 & ResNet-50 \cite{culane} & FastDraw Resnet50  \\
\hline\hline
Normal &  \textbf{87.4} & 85.9\\
Crowded & \textbf{64.1} & 63.6\\
Night & \textbf{60.6} & 57.8\\
No line & 38.1 & \textbf{40.6}\\
Shadow  & \textbf{60.7} & 59.9\\
Arrow & 79.0 & \textbf{79.4} \\
Dazzle & 54.1 & \textbf{57.0} \\
Curve  & 59.8 & \textbf{65.2}\\
Crossroad  & \textbf{2505} & 7013 \\
\hline
\end{tabular}
\end{center}
\caption{We compare FastDraw trained on CULane dataset to Resnet-50 on the CULane test set. We do not filter the lane predictions from FastDraw and achieve competitive results. While these scores are lower than those of SCNN \cite{culane}, we emphasize that architectural improvements such as those introduced in \cite{culane} are complementary to the performance of the FastDraw decoder.}
\label{tab:culane}
\end{table}

\subsection{Massachusetts}
\begin{figure}[t]
\begin{center}
   \includegraphics[width=0.98\linewidth]{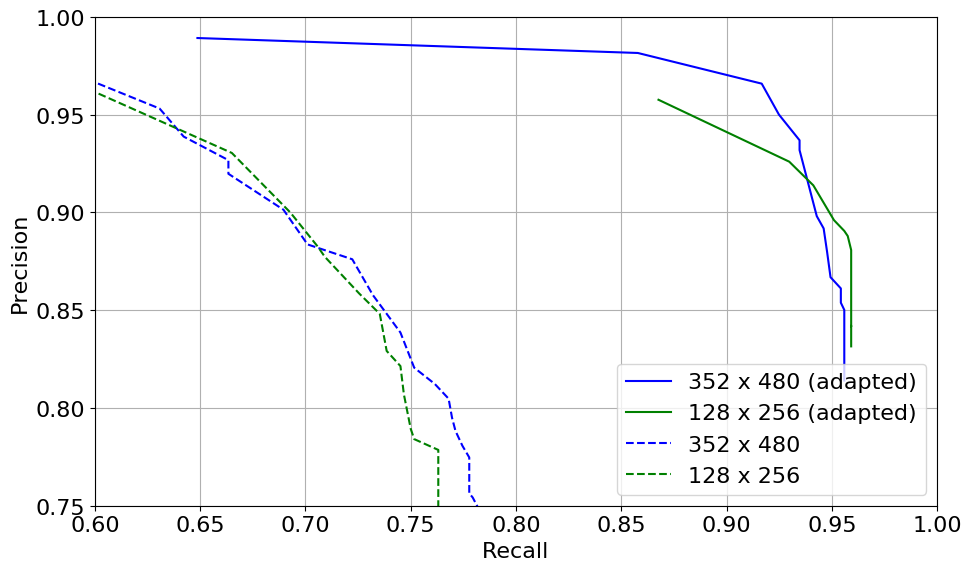}
\end{center}
   \caption{We label lane boundaries in a 300 image dataset of ``long tail" images to quantitatively evaluate the effect of the data augmentation. The precision/recall tradeoff for models trained on the augmented dataset is much better than the tradeoff of models trained without. The effect of the adaptation is demonstrated qualitatively in Figure \ref{fig:showoff}. }
\label{fig:ablation}
\end{figure}

We evaluate the ability of FastDraw to generalize to new scenes. In Figure \ref{fig:showoff} we demonstrate qualitatively that the network trained on style transferred training examples in addition to the Tusimple training examples can generalize well to night scenes, evening scenes, and rainy scenes. We emphasize that no additional human annotation was required to train FastDraw to be robust to these difficult environments.\\
\indent Additionally, we plot the precision/recall trade-off of FastDraw models trained with and without adaptation in Figure \ref{fig:ablation}. We use the same definition of false positive and false negative as used in the Tusimple evaluation. The augmentation compiles models that are markedly more robust to scene changes. We believe that these results echo recent findings that networks trained to do simple tasks on simple datasets learn low-level discriminative features that do not generalize \cite{textureinvar}. Unsupervised style transfer for data augmentation is offered as a naive but effective regularization of this phenomenon.\\

\section{Conclusion}
We demonstrate that it is possible to build an accurate model of lane detection that can adapt to difficult environments without requiring additional human annotation. The primary assumptions of our model is that lanes are curve segments that are functions of the height axis of an image and that a lane can be drawn iteratively by conditioning exclusively on the previous pixel that was determined to be part of the lane. With these assumptions, we achieve high accuracies on the lane detection task in standard and difficult environmental conditions.

{\small
\bibliographystyle{ieee}
\bibliography{egbib}
}

\end{document}